\documentclass[sigconf]{aamas} 
\usepackage{algorithm}
\usepackage{algpseudocode}
\DeclareMathOperator*{\argmin}{argmin} 
\DeclareMathOperator*{\argmax}{argmax} 
\newcommand\blfootnote[1]{%
  \begingroup
  \renewcommand\thefootnote{}\footnote{#1}%
  \addtocounter{footnote}{-1}%
  \endgroup
}
\usepackage{balance} % for balancing columns on the final page

\setcopyright{ifaamas}
\acmConference[AAMAS '21]{Proc.\@ of the 20th International Conference on Autonomous Agents and Multiagent Systems (AAMAS 2021)}{May 3--7, 2021}{Online}{U.~Endriss, A.~Now\'{e}, F.~Dignum, A.~Lomuscio (eds.)}
\copyrightyear{2021}
\acmYear{2021}
\acmDOI{}
\acmPrice{}
\acmISBN{}

\acmSubmissionID{94}

\title[AAMAS-2021 Formatting Instructions]{Cooperative and Competitive Biases 
\\for Multi-Agent Reinforcement Learning}

\author{Heechang Ryu}
\affiliation{
  \institution{KAIST}
  \city{Daejeon, Republic of Korea}}
\email{rhc93@kaist.ac.kr}

\author{Hayong Shin}
\affiliation{
  \institution{KAIST}
  \city{Daejeon, Republic of Korea}}
\email{hyshin@kaist.ac.kr}

\author{Jinkyoo Park\textsuperscript{$\ast$}}
\affiliation{
  \institution{KAIST}
  \city{Daejeon, Republic of Korea}}
\email{jinkyoo.park@kaist.ac.kr}

\begin{abstract}
Training a multi-agent reinforcement learning (MARL) algorithm is more challenging than training a single-agent reinforcement learning algorithm, because the result of a multi-agent task strongly depends on the complex interactions among agents and their interactions with a stochastic and dynamic environment. We propose an algorithm that boosts MARL training using the biased action information of other agents based on a friend-or-foe concept. For a cooperative and competitive environment, there are generally two groups of agents: cooperative-agents and competitive-agents. In the proposed algorithm, each agent updates its value function using its own action and the biased action information of other agents in the two groups. The biased joint action of cooperative agents is computed as the sum of their actual joint action and the imaginary cooperative joint action, by assuming all the cooperative agents jointly maximize the target agent's value function. The biased joint action of competitive agents can be computed similarly. Each agent then updates its own value function using the biased action information, resulting in a biased value function and corresponding biased policy. Subsequently, the biased policy of each agent is inevitably subjected to recommend an action to cooperate and compete with other agents, thereby introducing more active interactions among agents and enhancing the MARL policy learning. We empirically demonstrate that our algorithm outperforms existing algorithms in various mixed cooperative-competitive environments. Furthermore, the introduced biases gradually decrease as the training proceeds and the correction based on the imaginary assumption vanishes. 
\end{abstract}


\newcommand{\BibTeX}{\rm B\kern-.05em{\sc i\kern-.025em b}\kern-.08em\TeX}

\begin{document}

%%% The following commands remove the headers in your paper. For final 
%%% papers, these will be inserted during the pagination process.

\pagestyle{fancy}
\fancyhead{}

%%% The next command prints the information defined in the preamble.

\maketitle 

%%%%%%%%%%%%%%%%%%%%%%%%%%%%%%%%%%%%%%%%%%%%%%%%%%%%%%%%%%%%%%%%%%%%%%%%

\section{Introduction}

Reinforcement learning (RL) algorithms solve\blfootnote{\textsuperscript{$\ast$}Corresponding author} sequential decision-making problems using experiences obtained by a single agent (decision maker) dynamically interacting with an environment. The RL algorithms typically estimate an action-value function ($Q$-function) or a decision-making policy, using various function approximators (\emph{e.g.,} deep neural networks), to model how a particular action (decision) affects future outcomes. From this model, the optimal action in the current state for completing a target task can be deduced \cite{sutton2018reinforcement}. The success of RL algorithms for solving various tasks depends on how effectively they learn such temporal interactions between an action and a future outcome.

In particular, when the RL algorithm is trained with a sparse or delayed reward, \emph{i.e.,} a reward signal is infrequently realized after an agent executes action, it becomes difficult to estimate the $Q$-function or policy. This is because it is challenging for the RL agent to learn the dynamic causal effect of its action on future outcomes, owing to the sparse and delayed reward signal. A representative example of such a task with a sparse reward is a goal-oriented task, in which a binary reward is given only when an agent reaches a goal. To solve this task, HER \cite{andrychowicz2017hindsight} has been proposed to learn the $Q$-function or policy using the reward signals obtained from failed tasks (episodes), while considering these reward signals as being obtained from the successful tasks that are different from the original task. This strategy of HER transforms the sparse-reward environment into a dense-reward environment, enabling the RL agent to easily learn the $Q$-function or policy. In addition, HPG \cite{rauber2017hindsight} extends the concept of HER to efficiently generalize learning about different goals using information obtained by the current policy for a specific goal.

Recently, to enhance the performance of multi-agent reinforcement learning (MARL) algorithms, which are extensions of the RL algorithms to multi-agent settings, various strategies have been proposed. Some researchers have proposed an intrinsic reward to induce certain collective behaviors of multiple agents that are believed to help achieve the objective of a multi-agent task. For example, intrinsic reward is designed to promote the agents to execute actions influencing other agents' state transitions \cite{wang2019influence,bohmer2019exploration} or visit unexplored state spaces more frequently \cite{iqbal2019coordinated}. However, in general, designing a good intrinsic reward is difficult because it requires prior knowledge of the types of interactions that help solve the multi-agent task. Moreover, designing an effective intrinsic reward often requires an iterative reward-shaping procedure until satisfactory performance is reached. Consequently, for multi-agent tasks where a sparse reward is given and prior knowledge is not available, another effective method for MARL should be developed.

To address these difficulties, we herein propose an algorithm, called \textbf{F}riend-or-\textbf{F}oe multi-agent \textbf{D}eep \textbf{D}eterministic \textbf{P}olicy \textbf{G}radient (F2DDPG), which boosts MARL training using the biased action information of other agents based on a friend-or-foe concept \cite{friend-foe}. F2DDPG adopts an actor-critic algorithm in a centralized training and decentralized execution (CTDE) framework \cite{lowe2017multi}. For a cooperative and competitive environment, there are generally two groups of agents: an ally group composed of cooperative agents to a target agent and an enemy group composed of competitive agents to a target agent. In the proposed algorithm, each agent updates its $Q$-function (critic) using its own action and the biased action information of other agents in the two groups. The biased joint action of cooperative agents is computed as the sum of their actual joint action and the imaginary cooperative joint action obtained by assuming that all the cooperative agents jointly maximize the target agent's critic. The biased joint action of competitive agents can be computed similarly. Each agent then updates its own critic using biased action information, resulting in a biased critic and corresponding biased policy. Thereafter, the biased policy of each agent is inevitably subjected to recommend an action to cooperate and compete with other agents, which introduces more active interactions among agents, and thus, enhances the MARL policy learning.

Using biased actions in estimating a critic can be viewed as a different version of using the biased reward (intrinsic reward) in the estimation to induce the intended interactions among agents. However, we do not compare our method with MARL approaches using the intrinsic reward because they require a certain type of prior knowledge about the environment (game). Instead, we compare F2DDPG with M3DDPG \cite{li2019robust}, the latter of which uses modified (biased) action information of other agents when training the critic; it modifies the other agents' actions in an adversarial manner to induce a robust policy.
Empirically, we demonstrate that our algorithm outperforms existing algorithms in four mixed cooperative-competitive game scenarios, in which the agents have cooperative and competitive interactions \cite{lowe2017multi}. Furthermore, we empirically show that the introduced biases gradually decrease as the training proceeds and that the correction based on the imaginary assumption vanishes.

\section{Related Work}
\subsection{Friend-or-Foe Q-Learning}
In general-sum games, such as mixed cooperative-competitive games, friend-or-foe Q-learning (FFQ) \cite{friend-foe} has been proposed to provide strong convergence guarantees compared to the existing Nash-equilibrium-based learning rule \cite{hu1998multiagent}. It requires that other agents be identified as either `friend' (ally) or `foe' (enemy). FFQ then assumes that agent $i$'s friends are working together to maximize agent $i$'s value, while agent $i$'s foes are working together to minimize agent $i$'s value. Thus, $n$-player FFQ
considers any game as a two-player zero-sum game with an extended action set, and is easy to implement for multiple agents.

\subsection{Biases in RL and MARL}
In RL and MARL, various forms of inductive bias have been used to improve learning. The most straightforward inductive biases entail designing network structures for the critic or policy, such as attention networks \cite{iqbal2019actor}, graph neural networks \cite{ryu2020multi}, and implicit communication structures \cite{roy2019promoting}. However, biases in game information, such as state, reward, and action have also been used in an attempt to boost training. We review the biases in the information in this subsection.

\subsubsection{Biases in States.}
Bias has been reported to help train RL by injecting a biased belief regarding the state at the initialization stage of the $Q$-table \cite{hailu1999amount}. For example, in a goal-oriented task, if the goal is known in advance, biased information about the state, near and far from the goal, is injected into the $Q$-table before training. In addition, a distributed $Q$-learning algorithm for a cooperative multi-agent setting has been proposed based on the optimistic assumption \cite{lauer2000algorithm}. 
Under this assumption, the algorithm biasedly updates the $Q$-table only when the new value for $Q$ is greater than the current value in the current state. 

\subsubsection{Biases in Rewards.}
Intrinsic rewards have been proposed as a bias for multi-agent exploration to induce certain collective behaviors of agents, based on prior knowledge of the types of interactions that help solve the multi-agent task. The intrinsic rewards are provided when one agent's action affects the state transitions of other agents \cite{wang2019influence,bohmer2019exploration} and when all the agents explore only different or the same areas for the task of collecting scattered treasures \cite{iqbal2019coordinated}.

\subsubsection{Biases in Actions.} 

M3DDPG \cite{li2019robust} has been proposed to learn a robust policy using other agents' action information corrupted with adversarial noise. In this approach, each agent assumes that other agents provide the adversarial actions to the target agent, and updates its critic using such adversarial action information. To compute the adversarial action (biased action), each agent modifies the actions of other agents in the direction that minimizes the target agent's critic. M3DDPG asserts that the trained policy using this biased action information outperforms its baseline algorithm, MADDPG \cite{lowe2017multi}. However, the limitation of this approach is that it does not consider the relationships among agents. It assumes that all the agents are adversarial to the target agent, regardless of whether they are allies or enemies in the cooperative and competitive game; this assumption is inconsistent with the actual situation.

Notably, M3DDPG is similar to our method as it uses the biased information of other agents' actions. However, F2DDPG explicitly considers the roles of agents in a cooperative and competitive environment, in which the cooperative and competitive agents are known a priori. In addition, our method can be justified by the well-known FFQ \cite{friend-foe}. We mainly compare the performance of our method with that of M3DDPG, while not addressing other approaches using different types of information bias. In this study, we assume that it is possible to identify other agents as being either cooperative (allies) or competitive (enemies) in mixed cooperative-competitive environments, to fully implement and compare the proposed method with the baseline methods.

\begin{figure*}[t!]
    \centering
    \includegraphics[scale = 0.451]{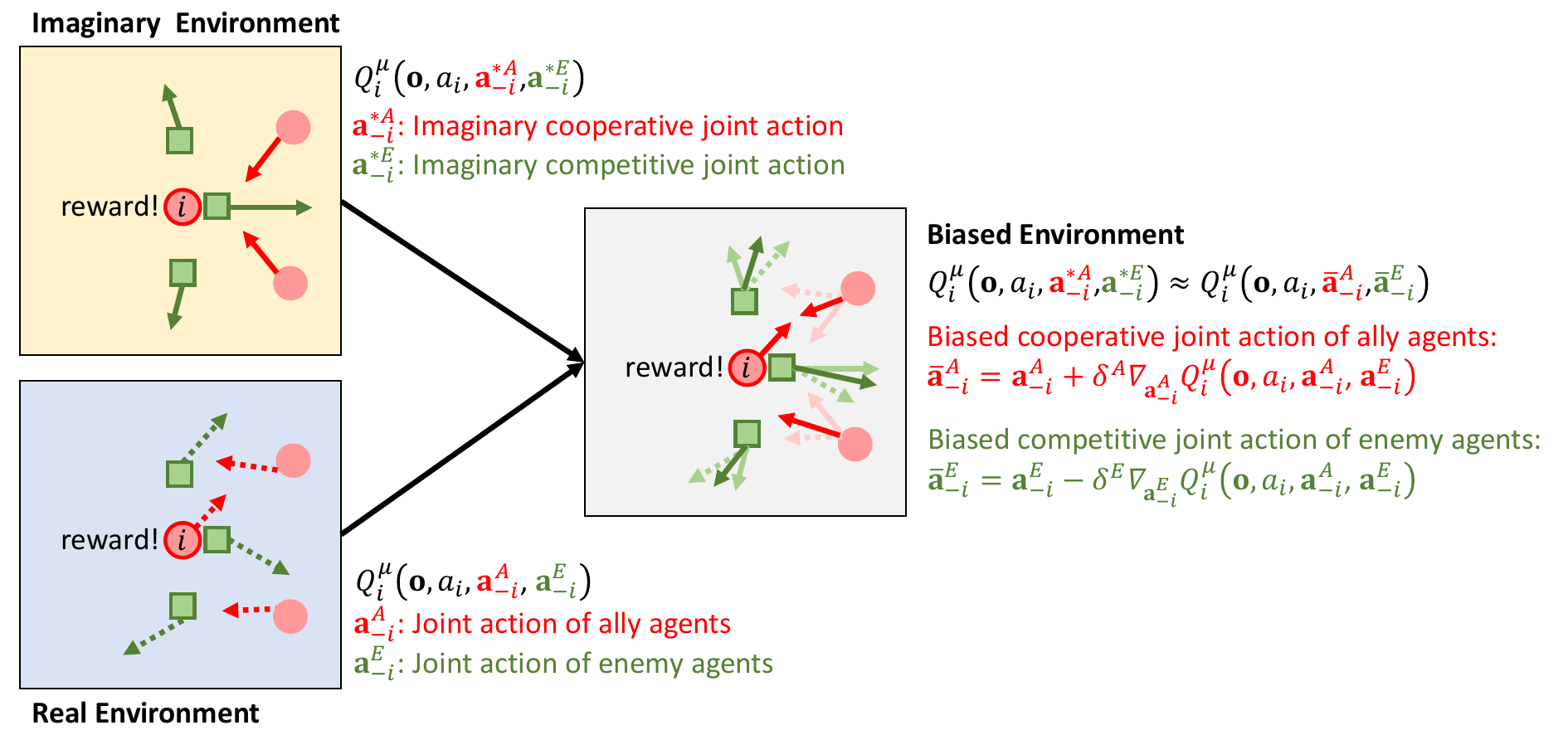}
    \caption{Overview of F2DDPG.}
    \label{fig:overview}
\end{figure*}

\section{Background}
\subsection{Partially Observable Markov Game} 
We consider a partially observable Markov game \cite{littman1994markov}, which is an extension of the partially observable Markov decision process to a game with multiple agents. A partially observable Markov game for $\mathnormal{N}$ agents is defined as follows: $s\in\mathcal{S}$ denotes the global state of the game; ${o_i}\in\mathcal{S}\mapsto\mathcal{O}_{i}$ denotes a local observation that agent $i$ can acquire correlated with the state; and $a_i\in\mathcal{A}_{i}$ is an action of agent $i$. The reward for agent $i$ is obtained as a function of state $s$ and joint action $\mathbf{a}$ as ${r}_{i}:\mathcal{S}\times\mathcal{A}_1\times\dots\times\mathcal{A}_N\mapsto\mathbb{R}$. The state evolves to the next state according to the state transition function $\mathcal{T}:\mathcal{S}\times\mathcal{A}_1\times\dots\times\mathcal{A}_N\mapsto\mathcal{S}$.
The initial state is determined by the initial state distribution $\rho:\mathcal{S}\mapsto[0,1]$.
Agent $i$ aims to maximize its discounted return $R_i=\sum_{t=0}^{T} \gamma^t r_{i,t}$, where $\gamma\in[0,1]$ is a discount factor.

\subsection{Multi-Agent Deep Deterministic Policy Gradient (MADDPG)} 
While the policy can be deterministic, $a=\mu(s)$, or stochastic, $a\sim\pi(\cdot\arrowvert s)$, deterministic policy gradient (DPG) \cite{silver2014deterministic} for RL adopts a deterministic policy. DPG aims to directly derive the deterministic policy, $a=\mu(s;\theta)$, that maximizes the expected return or objective $\mathcal{J(\theta)}=\mathbb{E}_{s\sim\rho^\mu,a\sim\mu_\theta}[R]\approx\mathbb{E}_{s\sim\rho^\mu,a\sim\mu_\theta}[Q^{\mu}(s,a;\phi)]$, where $Q^{\mu}(s,a;\phi)= \mathbb{E}_{s'}[r+\gamma\mathbb{E}_{a'\sim\mu_\theta}[Q^\mu(s',a';\phi)]]$. Parameter $\theta$ of $\mu(s;\theta)$ is subsequently optimized by the gradient of $\mathcal{J(\theta)}$ as $\nabla_\theta\mathcal{J}(\theta)=\mathbb{E}_{s\sim\mathcal{D}}[\nabla_\theta\mu(s;\theta){\nabla_a}Q^{\mu}(s,a;\phi)\arrowvert_{a=\mu(s;\theta)}]$. $\mathcal{D}$ is an experience replay buffer that stores $(s,a,r,{s'})$ samples obtained from the training episodes. Deep deterministic policy gradient (DDPG) \cite{lillicrap2015continuous}, an actor-critic algorithm based on DPG, uses deep neural networks to approximate critic $Q^{\mu}(s,a;\phi)$ and actor $\mu(s;\theta)$ of the agent.

MADDPG is a multi-agent extension of DDPG for deriving decentralized policies in the CTDE framework. In MADDPG, each agent learns an individual policy that maps the observation to its action to maximize its expected return, which is approximated by the $Q$-network. MADDPG comprises individual $Q$-networks and policy networks for each agent. MADDPG lets the $Q$-network (centralized critic) of agent $i$ be trained by minimizing the loss with the target $Q$-value, $y_i$, as follows: 
\begin{equation}
\begin{gathered} \label{eq:maddpg_Q_update}
\mathcal{L}(\phi_i)={\mathbb{E}_{\mathbf{o},\mathbf{a},r,{\mathbf{o}'}\sim \mathcal{D}}}[(Q_i^{\mu}(\mathbf{o},\mathbf{a};\phi_i)-y_i)^2], 
\\ y_i=r_i+\gamma {Q_i^{\mu'}}({\mathbf{o}'},\mathbf{a}';{\phi'_i})\arrowvert_{a_j'=\mu_j'(o'_j;\theta_j')},
\end{gathered}
\end{equation}
where $\mathbf{o}=(o_1,\dots,o_N)$ and $\mathbf{a}=(a_1,\dots,a_N)$ represent the joint observation and joint action of all agents, respectively. $\mathcal{D}$ is an experience replay buffer that stores $(\mathbf{o},\mathbf{a},r,{\mathbf{o}'})$ samples obtained from the training episodes. $Q^{\mu'}$ and $\mu'$ are target networks for the stable learning of $Q$ and policy networks. The policy network (actor), $\mu_i(o_i;\theta_i)$, of agent $i$ is optimized using the gradient computed as
\begin{equation} \label{eq:MADDPG_policy_update}
\nabla_{\theta_i}\mathcal{J}(\theta_i)=\mathbb{E}_{\mathbf{o},\mathbf{a}\sim \mathcal{D}}[\nabla_{\theta_i}\mu_i(o_i;\theta_i)\nabla_{a_i}Q_i^{\mu}(\mathbf{o},\mathbf{a};\phi_i )\arrowvert_{a_i=\mu_i(o_i;\theta_i)}].
\end{equation}

\section{Methods}
F2DDPG learns the critic and actor of each agent using the biased action information of other agents based on a friend-or-foe concept \cite{friend-foe}. In F2DDPG, as shown in Figure \ref{fig:overview}, each agent has two perceptions on the environment: 
\begin{itemize}
    \item The real environment, where agent $i$'s true critic can be estimated using its own action and the realized (actual) actions, $(a_i,\mathbf{a}^A_{-i}, \mathbf{a}^E_{-i})$;
    \item An imaginary environment, where agent $i$'s imaginary critic can be estimated using its own action and the imaginary cooperative and competitive joint actions, $(a_i,\mathbf{a}^{*A}_{-i}, \mathbf{a}^{*E}_{-i})$. These are computed based on the assumption that all the ally agents execute the cooperative joint action to agent $i$, while all the enemy agents execute the competitive joint action to agent $i$.
\end{itemize}

In F2DDPG, each agent learns the decentralized actor (policy) by applying the following three iterative procedures: (1) computing the biased actions, $\overline{\mathbf{a}}^A_{-i}$ and $\overline{\mathbf{a}}^E_{-i}$, by combining the real joint action and imaginary cooperative and competitive joint actions; (2) updating the biased critic using the biased actions; and (3) updating the actor using the biased critic. 

Thus, the updated policy with biased cooperative-competitive joint actions is more likely to recommend such cooperative and competitive actions to other agents, which introduces meaningful interactions among agents and enhances policy learning. Therefore, using the biased actions, F2DDPG can learn the level of cooperation and competition among agents in a sample-efficient manner. The biased actions become increasingly closer to the actually executed actions as the training proceeds, implying that the biases in the actors vanish.

Figure \ref{fig:overview} illustrates how agent $i$ (predator) updates its critic in F2DDPG, while playing the 3 vs. 3 predator-prey game. In this game, the three predators (red circles) attempt to capture the three prey (green squares) together. Because the prey are faster than the predators, the predators (ally group) must cooperate strategically to capture the prey (enemy group). In the early stages of learning, however, it is difficult for the predators to achieve a reward. Whenever the reward is realized, whether by chance or strategic moves, predator $i$ computes imaginary actions by assuming that the reward is realized when the other two predators choose the optimal cooperative joint action for predator $i$, while the other three prey execute the competitive joint action to predator $i$. Agent $i$ updates its critic using its own action and the one-step biased joint cooperative-competitive actions of other agents.

\begin{algorithm*}[t]
\caption{Friend-or-Foe Multi-Agent Deep Deterministic Policy Gradient Algorithm for $N$ agents}
\label{alg:F2DDPG}
\begin{algorithmic}[1]
\State Initialize actor networks $\mu$, critic networks $Q$, target networks $\mu'$ and $Q'$, experience replay buffer $\mathcal{D}$
\For {episode $= 1$ to $M$}
\State Initialize a random process $\mathcal{N}$ for action exploration
\State Receive initial observation $\mathbf{o}$
\For {$t = 1$ to $T$}
\State For each agent $i$, select action $a_i = \mu_i(o_i;\theta_i)+\mathcal{N}_t$
\State Execute actions $\mathbf{a}$ and receive reward $\mathbf{r}=(r_1,\dots,r_N)$ and new observation $\mathbf{o}'$
\State Store $(\mathbf{o},\mathbf{a},\mathbf{r},\mathbf{o}')$ in experience replay buffer $\mathcal{D}$
\State $\mathbf{o} \leftarrow \mathbf{o}'$
\For {agent $i=1$ to $N$}
\State Sample a random minibatch of $\mathcal{B}$ samples $(\mathbf{o}^b,\mathbf{a}^b,\mathbf{r}^b,\mathbf{o}'^b)$ from $\mathcal{D}$
\State Set $y^b_i=r^b_i+{\gamma}Q_i^{\mu'}({\mathbf{o}'^b},a'_i,\overline{\mathbf{a}}'^A_{-i},\overline{\mathbf{a}}'^E_{-i};\phi'_i)\arrowvert_{a_i'=\mu_i'(o_i'^b;\theta_i')}$,
\State $\overline{\mathbf{a}}'^A_{-i}=\mathbf{a}'^A_{-i}+\delta^{A}\nabla_{\mathbf{a}'^A_{-i}}Q_i^{\mu'}({\mathbf{o}'^b},a'_i,\mathbf{a}'^A_{-i},\mathbf{a}'^{E}_{-i};\phi_i')\arrowvert_{a_k'=\mu_k'(o'^b_k;\theta_k')}$,
%  for agent $j$ cooperative to agent $i$
\State $\overline{\mathbf{a}}'^E_{-i}=\mathbf{a}'^E_{-i}-\delta^{E}\nabla_{\mathbf{a}'^E_{-i}}Q_i^{\mu'}({\mathbf{o}'^b},a'_i,\mathbf{a}'^A_{-i},\mathbf{a}'^{E}_{-i};\phi_i')\arrowvert_{a_k'=\mu_k'(o'^b_k;\theta_k')}$
\State Update critic by minimizing the loss: $\mathcal{L}(\phi_i)=\frac{1}{\mathcal{B}} \sum_b (Q_i^{\mu}({\mathbf{o}^b},\mathbf{a}^b;\phi_i)-y^b_i)^2$
\State Update actor using the sampled policy gradient: $\nabla_{\theta_i}\mathcal{J}(\theta_i)
\approx \frac{1}{\mathcal{B}} \sum_b  \nabla_{\theta_i}\mu_i(o^b_i;\theta_i)\nabla_{a_i}Q_i^{\mu}({\mathbf{o}^b},a_i,\overline{\mathbf{a}}^A_{-i},\overline{\mathbf{a}}^E_{-i};\phi_i )\arrowvert_{a_i=\mu_i(o^b_i;\theta_i)}$,

\State $\overline{\mathbf{a}}^A_{-i}=\mathbf{a}^{bA}_{-i}+\delta^A\nabla_{\mathbf{a}^{bA}_{-i}}Q_i^{\mu}({\mathbf{o}^b},a^b_i,\mathbf{a}^{bA}_{-i},\mathbf{a}^{bE}_{-i};{\phi_i})$, 
\State $\overline{\mathbf{a}}^E_{-i}=\mathbf{a}^{bE}_{-i}-\delta^E\nabla_{\mathbf{a}^{bE}_{-i}}Q_i^{\mu}({\mathbf{o}^b},a^b_i,\mathbf{a}^{bA}_{-i},\mathbf{a}^{bE}_{-i};{\phi_i})$
\EndFor
\State Update target network parameters for each agent $i$: $\phi_i' \leftarrow \tau\phi_i+(1-\tau)\phi_i'$ and $\theta_i' \leftarrow \tau\theta_i+(1-\tau)\theta_i'$
\EndFor
\EndFor
\end{algorithmic}
\end{algorithm*}

\subsection{Computing Biased Cooperative-Competitive Actions}
Considering relationships to agent $i$, in the mixed cooperative-competitive environment, we categorize all agents, except $i$, as cooperative and competitive to agent $i$ as follows:
\begin{itemize}
    \item $A(i)$: set of agents cooperative to agent $i$ (ally group);
    \item $E(i)$: set of agents competitive to agent $i$ (enemy group);
    \item $\mathbf{a}^A_{-i}$: joint action of agents in $A(i)$;
    \item $\mathbf{a}^E_{-i}$: joint action of agents in $E(i)$.
\end{itemize}

If one assumes that the agents in $A(i)$ and $E(i)$ jointly maximize and minimize the critic of agent $i$, respectively, then agent $i$ can estimate its critic as follows:
\begin{equation} \label{eq:F2DDPG_imagine_maxminQ}
    \overline{Q}_i^\mu(\mathbf{o},a_i;\phi_i)=\max\limits_{\mathbf{a}^A_{-i}}\min\limits_{\mathbf{a}^E_{-i}}Q_i^\mu(\mathbf{o},a_i,\mathbf{a}^A_{-i},\mathbf{a}^E_{-i};\phi_i).
\end{equation}
This estimated critic is biased because each agent in $A(i)$ or $E(i)$ executes its action to only maximize its own individual critic similar to the process in MADDPG.

The two optimal joint actions $(\mathbf{a}^{*A}_{-i}$, $\mathbf{a}^{*E}_{-i})$ that achieve the \textit{maxmin} value in Equation~\ref{eq:F2DDPG_imagine_argmaxminQ} are called a saddle-point equilibrium strategy, which is equivalent to the Nash equilibrium strategy, which satisfies
\begin{equation} \label{eq:F2DDPG_imagine_argmaxminQ}
\begin{gathered}
    \mathbf{a}^{*A}_{-i}=\argmax\limits_{\mathbf{a}^A_{-i}}Q_i^\mu(\mathbf{o},a_i,\mathbf{a}^A_{-i},\mathbf{a}^{*E}_{-i};\phi_i),
    \\\mathbf{a}^{*E}_{-i}=\argmin\limits_{\mathbf{a}^E_{-i}}Q_i^\mu(\mathbf{o},a_i,\mathbf{a}^{*A}_{-i},\mathbf{a}^{E}_{-i};\phi_i).
\end{gathered}
\end{equation}
We refer to these two actions in Equation~\ref{eq:F2DDPG_imagine_argmaxminQ} as the imaginary cooperative and competitive joint actions because these two joint actions rarely occur in the real environment.

The estimated critic of agent $i$ using Equation~\ref{eq:F2DDPG_imagine_argmaxminQ} can be used in a decentralized training setting, where each agent cannot observe other agents' actions during training, and thus, has to infer them. However, the current study focuses on developing an efficient MARL algorithm in the CTDE framework that allows each agent to observe other agents' actions during training. To help stabilize training without introducing any bias, other agents' actions are explicitly used when learning the critic and associated actor.

In this study, we propose to combine these two different learning paradigms to: (1) achieve reliable and stable learning using the true action information in the CTDE framework and (2) infuse the desirable behaviors (information biases) into agents using imaginary joint cooperative-competitive actions (biased action information) computed in the decentralized learning framework. To achieve both objectives, the proposed method computes the biased actions by combining the actual and imaginary actions as follows:
\begin{equation} \label{eq:F2DDPG_bias}
\begin{gathered}
    \overline{\mathbf{a}}^A_{-i}=\mathbf{a}^A_{-i}+\delta^A\nabla_{\mathbf{a}^A_{-i}}Q_i^{\mu}({\mathbf{o}},a_i,\mathbf{a}^A_{-i},\mathbf{a}^{E}_{-i};{\phi_i}),
    \\\overline{\mathbf{a}}^E_{-i}=\mathbf{a}^E_{-i}-\delta^E\nabla_{\mathbf{a}^E_{-i}}Q_i^{\mu}({\mathbf{o}},a_i,\mathbf{a}^A_{-i},\mathbf{a}^{E}_{-i};{\phi_i}).
\end{gathered}
\end{equation}
In Equation~\ref{eq:F2DDPG_bias}, we compute the one-step-biased cooperative-competitive joint actions, which approximate the imaginary cooperative-competitive joint actions in Equation~\ref{eq:F2DDPG_imagine_argmaxminQ}, using the partial gradient of critic $Q_i^{\mu}$. In addition, $\delta^A$ and $\delta^E$ are the step sizes for the biased cooperative and competitive joint actions, respectively. 

Note that computing $\overline{\mathbf{a}}^A_{-i}$ and $\overline{\mathbf{a}}^E_{-i}$ is computationally tractable because it only requires computing the partial gradient of the $Q$-network with respect to the joint action variables. In addition, $\delta^A$ and $\delta^E$ adjust the level of biases (injected cooperative and competitive biases). The optimal $\delta^A$ and $\delta^E$ can be empirically determined during training; however, these values do not significantly affect the learning performance because the partial gradient eventually becomes small, making the amount of action modification negligible.

\subsection{Learning the Biased Critic Using Biased Actions}
The biased actions in Equation \ref{eq:F2DDPG_bias} are then used to update the $Q$-network as:
\begin{equation}
\begin{gathered} \label{eq:F2DDPG_gradient_Q_update}
\mathcal{L}(\phi_i)={\mathbb{E}_{\mathbf{o},\mathbf{a},r,{\mathbf{o}'}\sim \mathcal{D}}}[(Q_i^{\mu}(\mathbf{o},\mathbf{a};\phi_i)-y_i)^2, 
\\ y_i=r_i+\gamma {Q_i^{\mu'}}({\mathbf{o}'},a'_i,\overline{\mathbf{a}}'^A_{-i},\overline{\mathbf{a}}'^E_{-i};\phi_i')\arrowvert_{a_i'=\mu_i'(o'_i;\theta_i')}],
\end{gathered}
\end{equation}
where $\overline{\mathbf{a}}'^A_{-i}$ and $\overline{\mathbf{a}}'^E_{-i}$ are, respectively, the biased actions computed using Equation~\ref{eq:F2DDPG_bias} with the target $Q$-network, $Q_i^{\mu'}$. The target network is designed to stabilize the learning of the $Q$-network by slowly changing the parameters $\phi'$ of the target $Q$-network \cite{lowe2017multi}.

\subsection{Learning the Actor from the Biased Critic}
The biased critic $Q_i^{\mu}$ is used to update the decentralized actor as a deterministic policy. The deterministic policy network $\mu_i$ is updated using the gradient computed as
\begin{equation}
\begin{gathered}
\label{eq:F2DDPG_policy_update}
\nabla_{\theta_i}\mathcal{J}(\theta_i)
=\qquad\qquad\qquad\qquad\qquad\qquad\qquad\qquad\qquad\\
\mathbb{E}_{\mathbf{o},\mathbf{a}\sim \mathcal{D}}[\nabla_{\theta_i}\mu_i(o_i;\theta_i)\nabla_{a_i}Q_i^{\mu}(\mathbf{o},a_i,\overline{\mathbf{a}}^A_{-i},\overline{\mathbf{a}}^E_{-i};\phi_i )\arrowvert_{a_i=\mu_i(o_i;\theta_i)}],
\end{gathered}
\end{equation} 
where $\overline{\mathbf{a}}^A_{-i}$ and $\overline{\mathbf{a}}^E_{-i}$ are the biased joint actions of the ally and energy agents, respectively, each of which is computed using Equation~\ref{eq:F2DDPG_bias} with the $Q$-network, $Q_i^{\mu}$. We estimate the gradient reliably with the biased actions computed using the sampled joint actions (executed true actions) from the experience reply buffer.

Each agent then updates the parameters of its own policy network using the computed policy gradient in Equation~\ref{eq:F2DDPG_policy_update}. Owing to the biased action information, the trained policy is also biased such that it is inevitably subjected to recommend an action to cooperate and compete with other agents. Consequently, the biased policy introduces more active interactions among agents, and thus, enhances the MARL policy learning. In addition, if the critic's gradient (bias) in the biased action and actual action from the learned policy gradually become similar, the introduced bias then naturally decreases, and the training eventually becomes unbiased, as the biased and actual actions become the same. For completeness, we provide the F2DDPG algorithm in Algorithm \ref{alg:F2DDPG}.

\setcounter{figure}{-3}
\begin{figure*}[t]
\begin{minipage}[c]{.24\textwidth}
  \centering
  \includegraphics[width=0.9\textwidth]{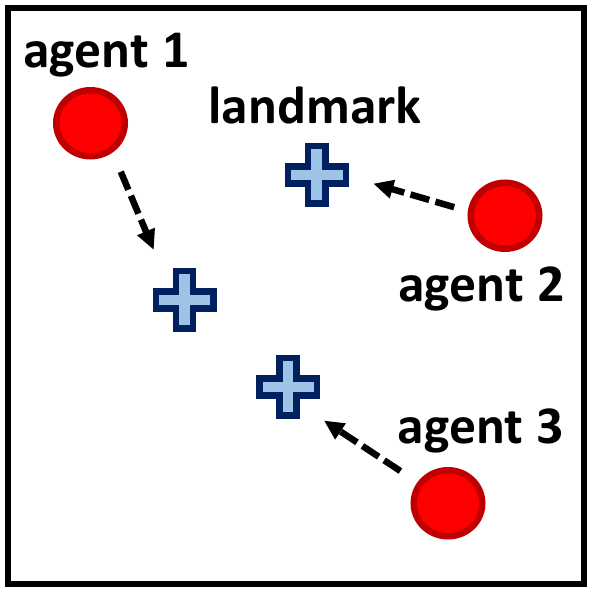}
  \captionsetup{labelformat=empty}
  \caption{a) Cooperative Navigation}
\end{minipage}
% \hspace{0.0cm}
\begin{minipage}[c]{.26\textwidth}
  \centering
  \includegraphics[width=0.83\textwidth]{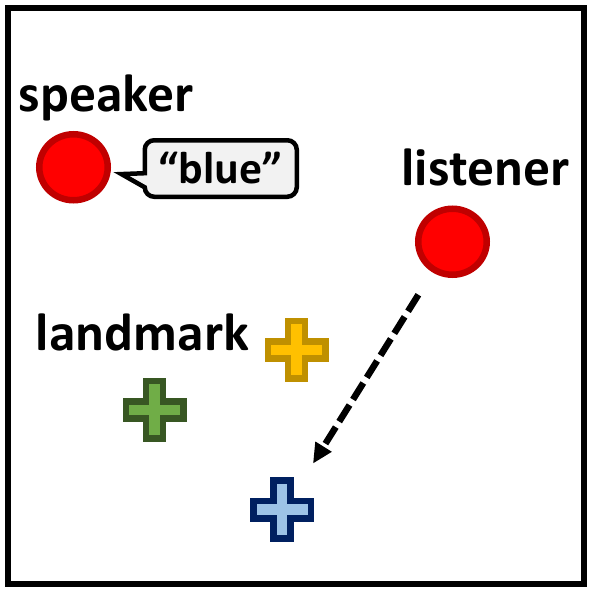}
  \captionsetup{labelformat=empty}
  \caption{b) Cooperative Communication}
\end{minipage}
% \hspace{0.0cm}
\begin{minipage}[c]{.24\textwidth}
  \centering
  \includegraphics[width=0.9\textwidth]{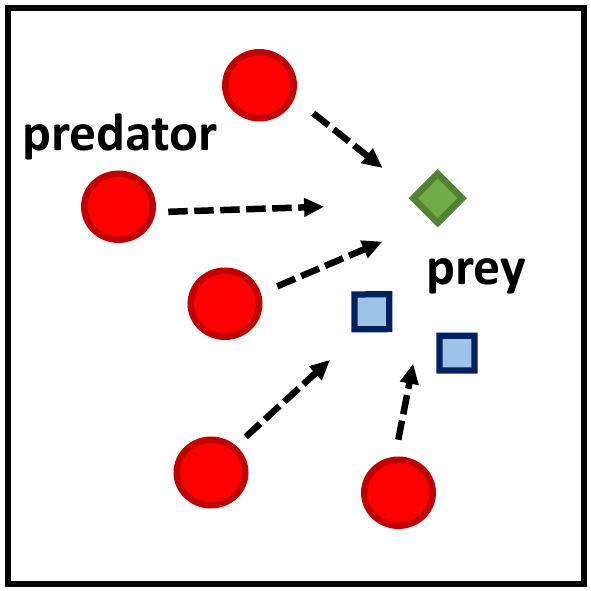}
  \captionsetup{labelformat=empty}
  \caption{c) Predator-Prey}
\end{minipage}
\begin{minipage}[c]{.24\textwidth}
  \centering
  \includegraphics[width=0.9\textwidth]{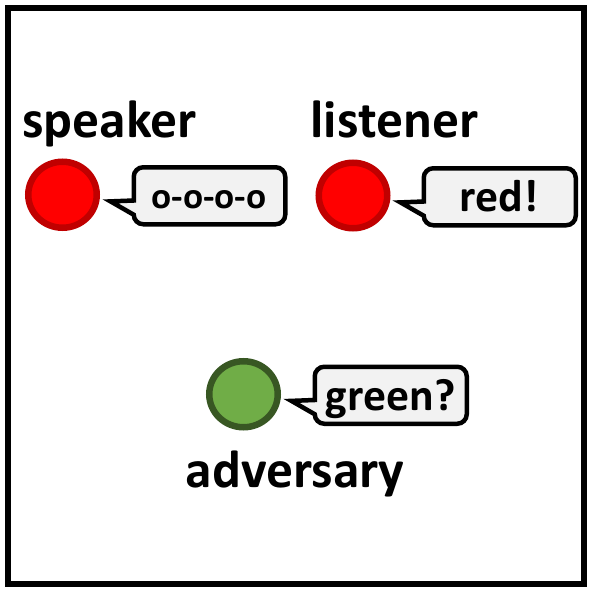}
  \captionsetup{labelformat=empty}
  \caption{d) Covert Communication}
\end{minipage}
\caption{Illustrations of the experimental environments.}
\label{fig:fig_illustrations}
\end{figure*}

In some cases of training F2DDPG, the critic's gradient (bias) dominates the actual action in the biased action in learning the $Q$-network and policy network. Thus, the magnitude of the gradient is made equal to the magnitude of the actual action to prevent the biased action from being extremely biased and destabilizing the training, as follows:
\begin{equation}
\begin{gathered} \label{eq:gradient_norm}
g=\nabla_{a_{-i}}Q_i^{\mu}({\mathbf{o}},\mathbf{a};{\phi_i}),\\
\overline{a}_{-i}=a_{-i}\pm\delta \lVert a_{-i}\rVert_2\frac{g}{\lVert g \rVert_2}.
\end{gathered}
\end{equation}
% This is a trick similar to the one used in \cite{li2019robust} to stabilize learning. 
The trick of Equation~\ref{eq:gradient_norm} is utilized in Equation~\ref{eq:F2DDPG_bias}.

\setcounter{figure}{-2}
\begin{figure*}[t]
\begin{minipage}[c]{.24\textwidth}
  \centering
  \includegraphics[width=1.02\textwidth]{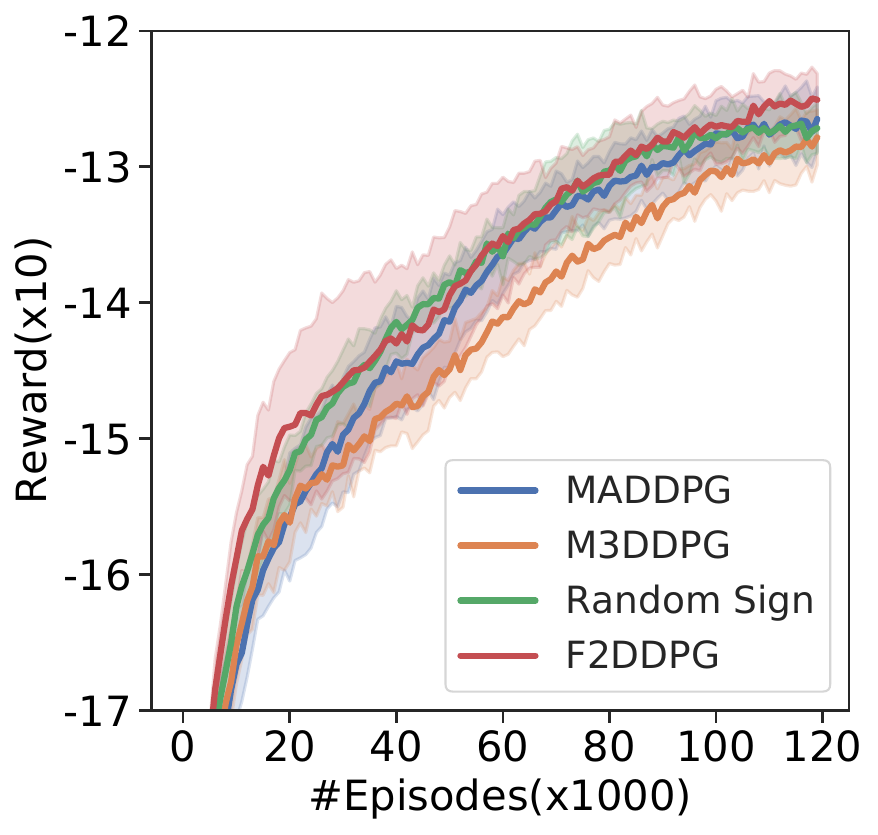}
  \captionsetup{labelformat=empty}
  \caption{a) Cooperative Navigation}
\end{minipage}
% \hspace{0.0cm}
\begin{minipage}[c]{.26\textwidth}
  \centering
  \includegraphics[width=0.95\textwidth]{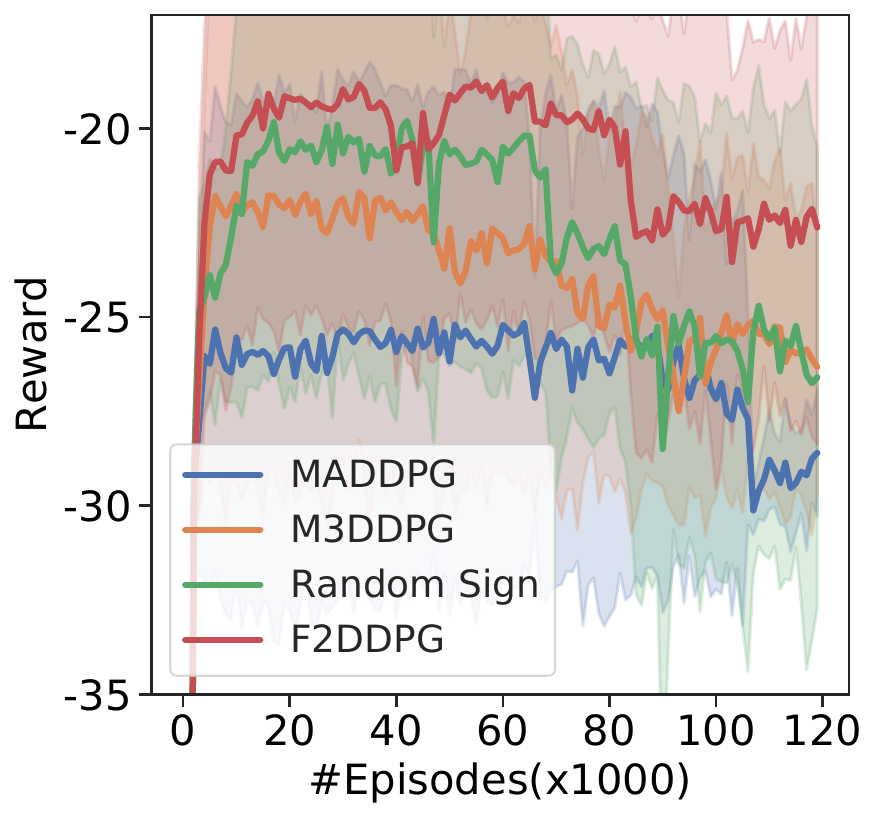}
  \captionsetup{labelformat=empty}
  \caption{b) Cooperative Communication}
\end{minipage}
% \hspace{0.0cm}
\begin{minipage}[c]{.24\textwidth}
  \centering
  \includegraphics[width=1.03\textwidth]{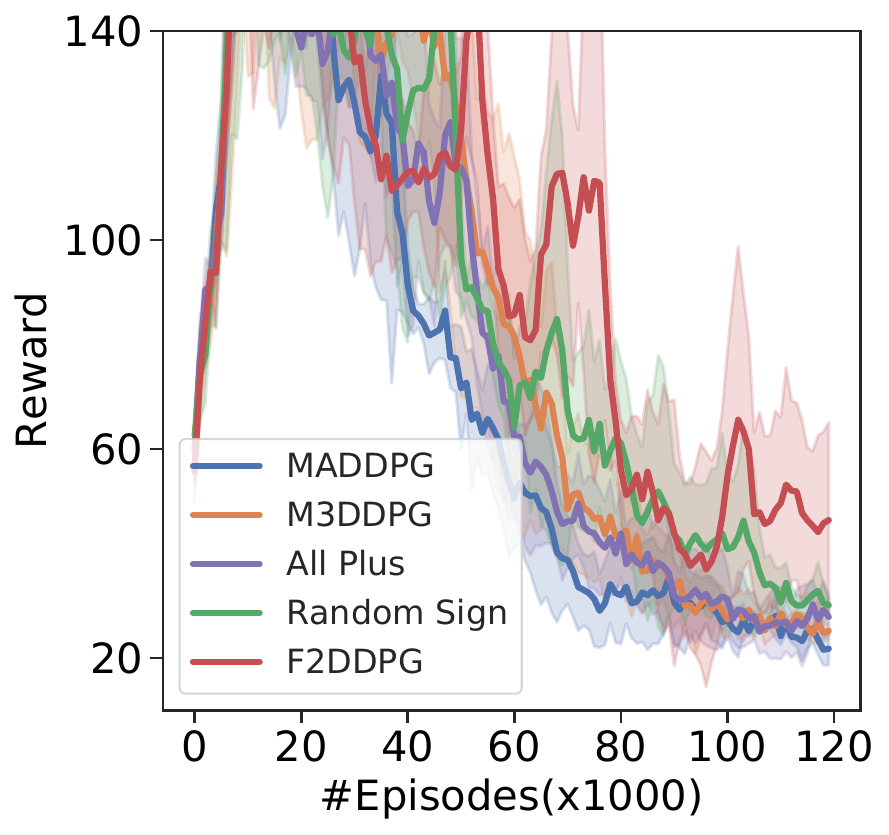}
  \captionsetup{labelformat=empty}
  \caption{c) Predator-Prey}
\end{minipage}
\begin{minipage}[c]{.24\textwidth}
  \centering
  \includegraphics[width=1\textwidth]{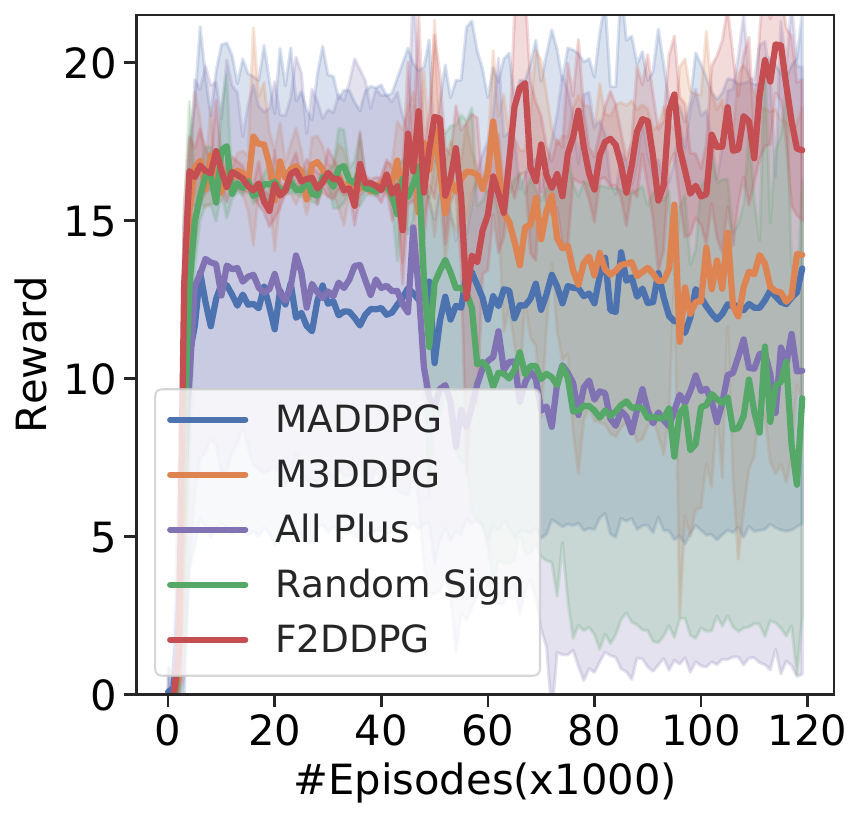}
  \captionsetup{labelformat=empty}
  \caption{d) Covert Communication}
\end{minipage}
\caption{ Rewards of the red agents in the experimental environments.}
\label{fig:results_env}
\end{figure*}

\section{Experiments}
Figure \ref{fig:fig_illustrations} shows the environments (games) used to evaluate the performances of the proposed and baseline algorithms. The environments are those used in previous studies \cite{lowe2017multi,li2019robust} and those designed to make MARL training more difficult. We assume that the agents in the environments observe the relative positions and velocities of all agents. In Figure \ref{fig:fig_illustrations}, games (a) and (b) correspond to cooperative environments with only agents in a cooperative relationship, and games (c) and (d) correspond to mixed cooperative-competitive environments with both cooperative and competitive agents. In games (b) and (d), the agents need to communicate with other agents depending on the purpose of the games. 

For the experiments in these games, we compare the performance of F2DDPG to the following baseline algorithms:
\begin{itemize}
    \item \textbf{MADDPG} \cite{lowe2017multi} is the algorithm that learns the critics and actors using only the actual action information in the CTDE framework.
    \item \textbf{M3DDPG} \cite{li2019robust} is the algorithm based on MADDPG. Instead of using the actual action information, this algorithm uses the noisy actions of other agents to update the critics and actors. In particular, this algorithm computes the adversarial noise for other agents' actions such that the noisy actions collectively minimize the target agent's critic. Owing to the use of adversarial noise, this algorithm is referred to as robust MARL. In the view of F2DDPG, this algorithm can be interpreted as one in which the competitive roles are infused, as an information bias, to all the agents, regardless of their roles 
    (cooperative or competitive)
    with respect to the target agent. 
    \item \textbf{All Plus} is the algorithm with only cooperative biases employed in the actions of other agents; other agents are assumed to maximize the target agent's critic, regardless of their roles.
    \item \textbf{Random Sign} is the algorithm with random biases employed in the actions of other agents. Other randomly selected agents are assumed to maximize the target agent's critic, while the remaining agents are assumed to minimize the target agent's critic, regardless of their roles. 
\end{itemize}
We can differentiate the proposed F2DDPG and other baseline algorithms depending on the type of information bias infused into other agents' actions. While the baseline algorithms either do not use any information bias (MADDPG) or use a certain information bias, regardless of the relationships among agents, F2DDPG is the only algorithm that aligns the information bias with the actual roles of agents. 
Note that we exclude the All Plus algorithm from the baseline algorithms in cooperative environments, such as games (a) and (b), because it has the same cooperative biases as F2DDPG for all the cooperative agents. In this study, all the performances are obtained by the trained policies with four different random seeds.

\subsection{Cooperative Navigation}
Cooperative navigation is a cooperative environment, as shown in Figure \ref{fig:fig_illustrations} (a), in which three cooperative agents (red circles) must reach three landmarks (blue crosses) without colliding with each other, while covering all of the landmarks. Every episode starts with randomly initialized positions for the agents and landmarks. The agents are collectively rewarded based on the distance of the nearest agent to each landmark and penalized for collisions with other agents during navigation. Thus, each agent must cooperate to occupy a distinct landmark without colliding with other agents.

As shown in Figure \ref{fig:results_env} (a), F2DDPG outperforms other baseline algorithms with faster training speed and higher converged rewards. We consider that the performance improvement of F2DDPG is a result of the optimal use of information bias corresponding to the agents' roles. F2DDPG constantly updates the critic and actor of each agent using biased joint actions, which induces agents' policies to recommend more coherently exploratory actions, especially toward cooperation. We believe that such coherent exploration helps the MARL policy learning more than random exploration.

In contrast, M3DDPG exhibits slow training in inducing cooperation among the three agents by learning only the adversarial action information of other agents, although the agents are in a cooperative relationship. Training with adversarial action information may enable robust policy learning; however, it is believed to be unhelpful in inducing cooperation among the agents. The Random Sign algorithm randomly injects cooperative or competitive biases for biased actions at every step of training, which leads to random noise in training. This random noise allows the algorithm to train faster than MADDPG and M3DDPG; however, the algorithm exhibits slower training than F2DDPG. 

\setcounter{figure}{4}
\begin{figure*}[t]
\begin{minipage}[c]{.145\textwidth}
  \centering
  \includegraphics[width=1.05\textwidth]{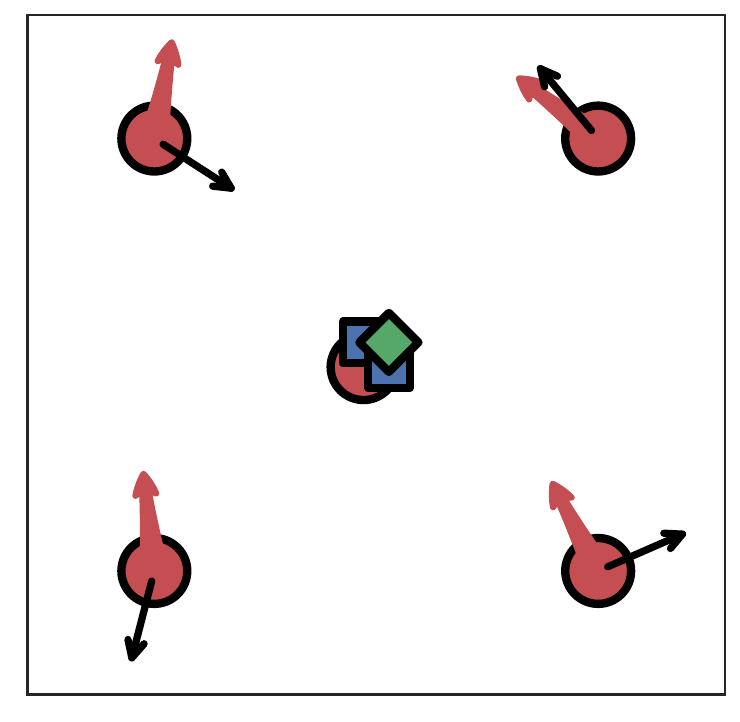}
\end{minipage}
\hspace{0.0cm}
\begin{minipage}[c]{.145\textwidth}
  \centering
  \includegraphics[width=1.05\textwidth]{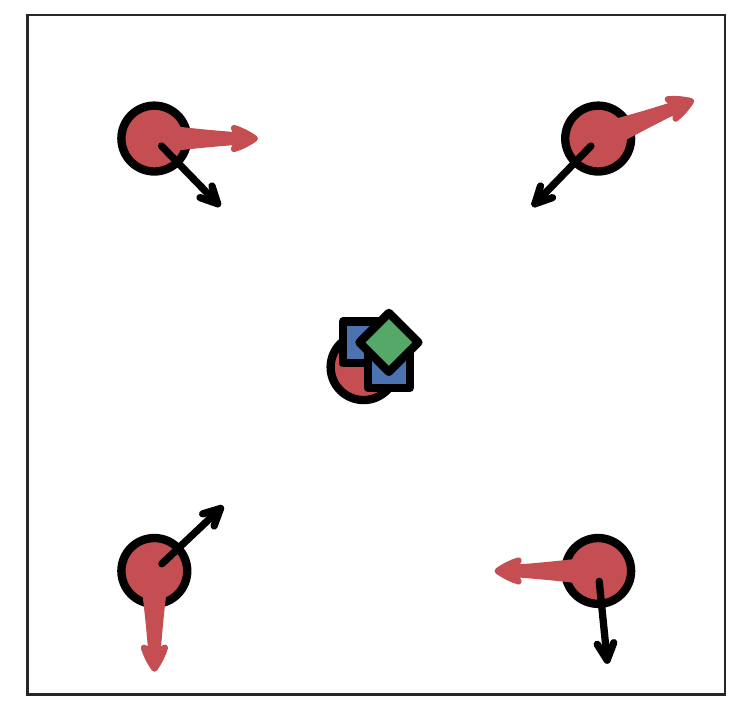}
\end{minipage}
\hspace{0.0cm}
\begin{minipage}[c]{.145\textwidth}
  \centering
  \includegraphics[width=1.05\textwidth]{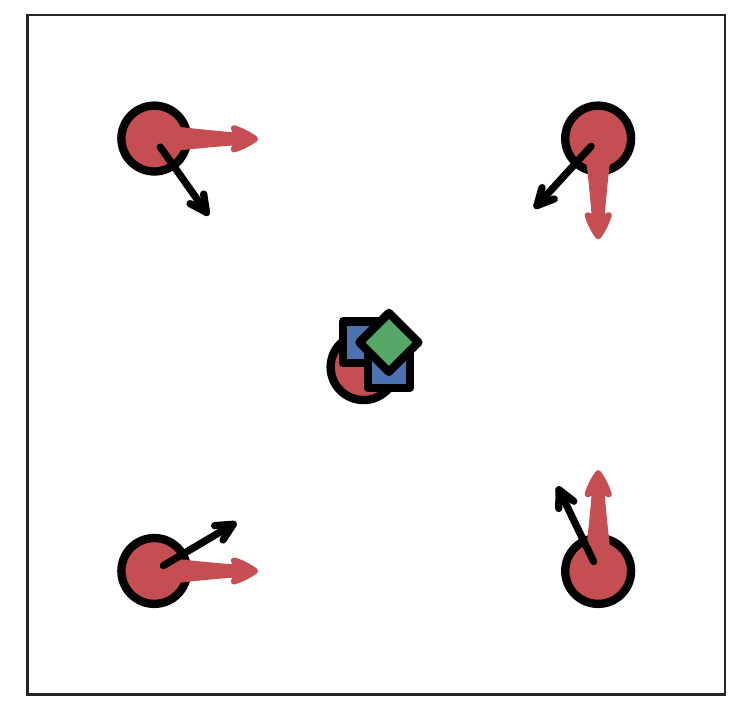}
\end{minipage}
\hspace{0.0cm}
\begin{minipage}[c]{.145\textwidth}
  \centering
  \includegraphics[width=1.05\textwidth]{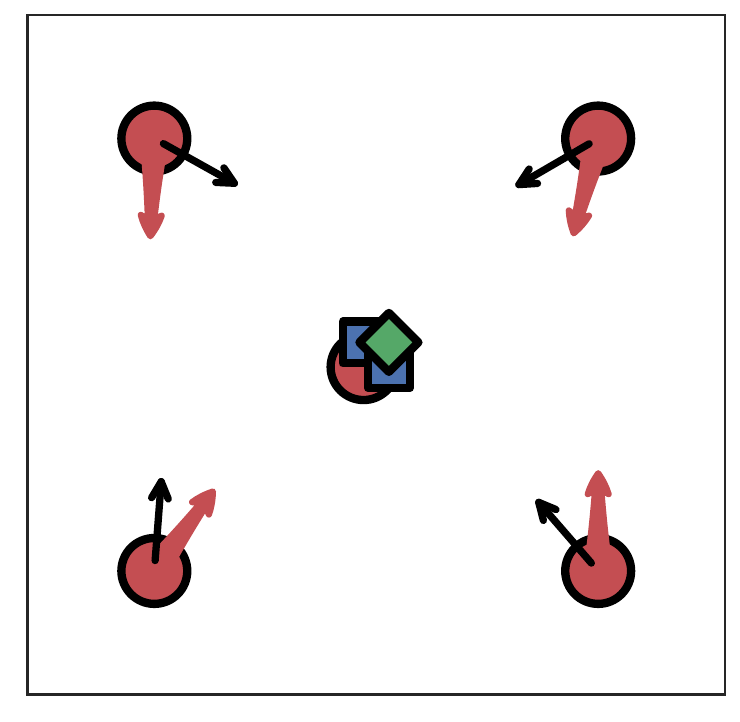}
\end{minipage}
\hspace{0.0cm}
\begin{minipage}[c]{.145\textwidth}
  \centering
  \includegraphics[width=1.05\textwidth]{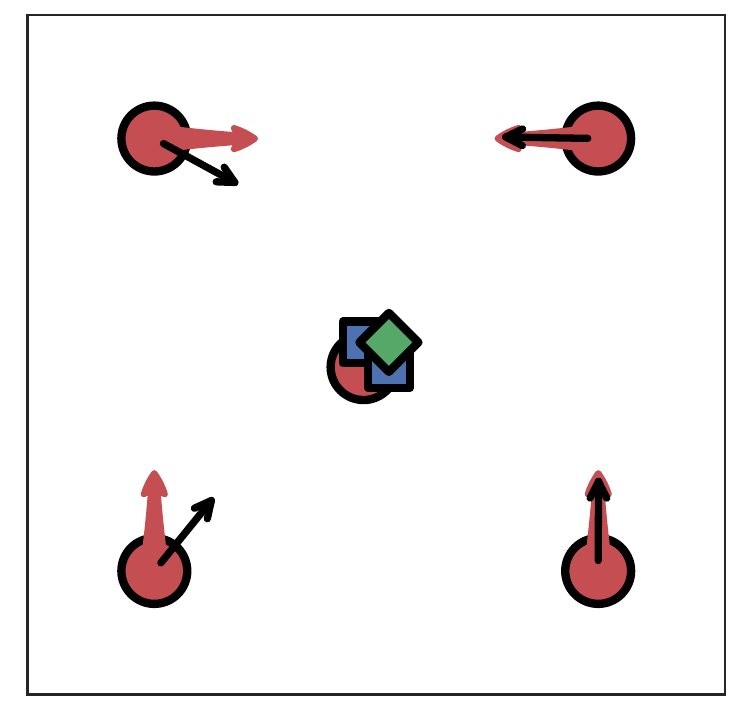}
\end{minipage}
\hspace{0.0cm}
\begin{minipage}[c]{.145\textwidth}
  \centering
  \includegraphics[width=1.05\textwidth]{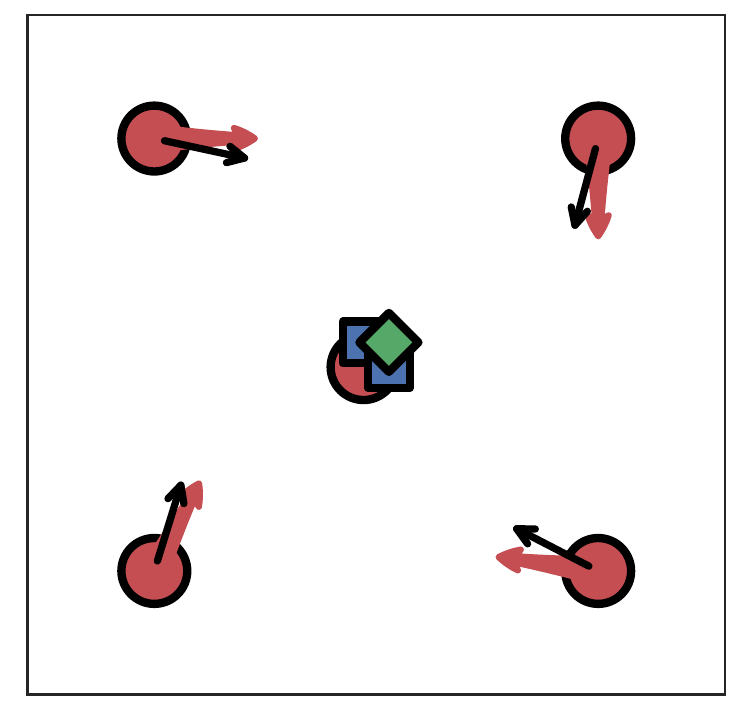}
\end{minipage}

\begin{minipage}[c]{.99\textwidth}
  \centering
  \includegraphics[width=0.93\textwidth]{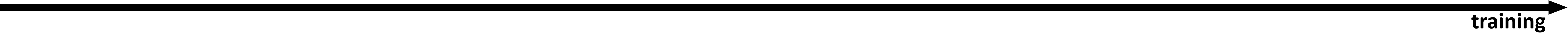}
\end{minipage}
\caption{Actual actions (red arrows) and biases (black arrows) for the centered predator as training proceeds.}
\label{fig:fixed_position_action}
\end{figure*}

\subsection{Cooperative Communication}
Cooperative communication is a cooperative environment, as shown in Figure \ref{fig:fig_illustrations} (b), with two cooperative agents, a speaker and listener (red circles), and three landmarks of differing colors. The listener must navigate to a landmark of a particular color. However, the listener does not know the landmark to which it must navigate, while observing the relative position and color of the landmarks. In contrast, the speaker observes the correct color of the landmark to which the listener must navigate, and broadcasts a message (communication vector) at each time step, which is observed by the listener. For each episode, the positions of the listener and landmarks are randomly initialized. The listener and speaker are rewarded based on listener's distance to the correct landmark. Thus, the speaker must learn to generate a message that optimally guides the listener to reach the correct landmark, and simultaneously, the listener must learn to decipher the message that is transmitted from the speaker and navigate to the correct landmark.

As shown in Figure \ref{fig:results_env} (b), F2DDPG outperforms other baseline algorithms with faster training speed and higher rewards. Additionally, M3DDPG exhibits better performance than MADDPG. Moreover, the Random Sign algorithm learns faster than MADDPG, possibly because of the enhanced exploration with random noises added to the observed actions. 

\begin{table*}[t]
  \caption{Fraction between the number of successful episodes and the total number of testing episodes in predator-prey games.}
  \label{table:results_predator-prey}
  \centering
    \begin{tabular}{ccccccc}
        \toprule
        \multicolumn{1}{c}{} &  \multicolumn{2}{c}{3 vs. 1} & \multicolumn{2}{c}{5 vs. 3} &
        \multicolumn{2}{c}{7 vs. 3}\\
        \cmidrule(r){2-3}
        \cmidrule(r){4-5}
        \cmidrule(r){6-7}
        \multicolumn{1}{c}{} & \multicolumn{1}{c}{$N_c\geq1$}&
        \multicolumn{1}{c}{$N_c\geq3$}&
        \multicolumn{1}{c}{$N_c\geq1$}&
        \multicolumn{1}{c}{$N_c\geq3$}&
        \multicolumn{1}{c}{$N_c\geq1$}&
        \multicolumn{1}{c}{$N_c\geq3$}\\
        \midrule
        \multicolumn{1}{c}{MADDPG} &2.75\scriptsize $\pm$ 0.83 &\textbf{0.25}\scriptsize $\pm$ 0.43 & 14.50\scriptsize $\pm$ 1.50 &1.25\scriptsize $\pm$ 1.09 &56.25\scriptsize $\pm$ 6.46 & 17.00\scriptsize $\pm$ 6.09\\
        \multicolumn{1}{c}{M3DDPG} & \textbf{3.75}\scriptsize $\pm$ 0.43 &\textbf{0.25}\scriptsize $\pm$ 0.43 & 14.25\scriptsize $\pm$ 0.43 & 1.25\scriptsize $\pm$ 0.83 & 55.75\scriptsize $\pm$ 6.17 & 13.25\scriptsize $\pm$ 3.63\\
        \multicolumn{1}{c}{All Plus} & \textbf{3.75}\scriptsize $\pm$ 1.09 & \textbf{0.25}\scriptsize $\pm$ 0.43 &15.75\scriptsize $\pm$ 2.28& 1.00\scriptsize $\pm$ 1.00 &62.75\scriptsize $\pm$ 7.39 & 14.24\scriptsize $\pm$ 4.14\\
         \multicolumn{1}{c}{Random Sign}  &1.50\scriptsize $\pm$ 0.50& 0.00\scriptsize $\pm$ 0.00 & 18.75\scriptsize $\pm$ 6.98 & 1.75\scriptsize $\pm$ 0.83 & 62.50\scriptsize $\pm$ 11.71& 19.25\scriptsize $\pm$ 6.37\\
         \multicolumn{1}{c}{F2DDPG}& 3.25\scriptsize $\pm$ 0.43 & \textbf{0.25}\scriptsize $\pm$ 0.43 &\textbf{32.50}\scriptsize $\pm$ 11.39 & \textbf{4.50}\scriptsize $\pm$ 2.29&
         \textbf{71.75}\scriptsize $\pm$ 7.52 & \textbf{28.75}\scriptsize $\pm$ 6.17\\
        \bottomrule
    \end{tabular}
\end{table*}

\subsection{Predator-Prey}
Predator-prey is a mixed cooperative-competitive environment, as shown in Figure \ref{fig:fig_illustrations} (c), in which the five predator agents (red circles) seek to capture the three prey agents (blue squares and green diamond), which is called 5 vs. 3 predator-prey. If there are $m$ predators and $n$ prey, it is denoted as $m$ vs. $n$ predator-prey. Because prey can move at a higher speed and have greater acceleration than predators, predators must cooperate to capture the prey. In particular, the green prey (green diamond) can move faster with greater acceleration than the blue prey (blue squares); the green and blue prey are factors of 3 and 1.3 faster than the predators, respectively. The positions of the predators and prey are randomly initialized for every episode. 

Each time the predators collide with (\emph{i.e.,} capture) the prey, the predators are collectively rewarded, while the prey are penalized. When the predators capture green prey, they are rewarded with a factor of 10 more than when they capture blue prey. The predators can capture prey multiple times during an episode. In the predator-prey game, the prey are trained with MADDPG, while the predators are trained with F2DDPG and other baseline algorithms.

\setcounter{figure}{1}
\begin{figure}[h!]
\begin{minipage}[t]{.235\textwidth}
  \centering
  \includegraphics[width=1.02\textwidth]{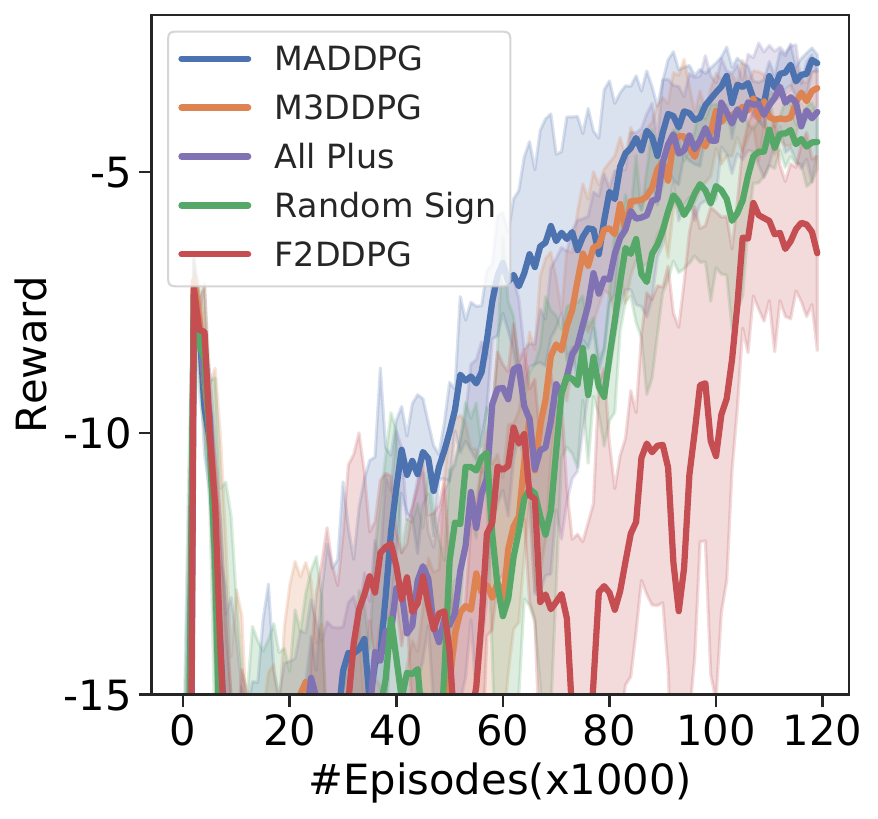}
  \captionsetup{labelformat=empty}
  \caption{a) Rewards of green prey}
\end{minipage}
% \hspace{0.0cm}
\begin{minipage}[t]{.235\textwidth}
  \centering
  \includegraphics[width=1.02\textwidth]{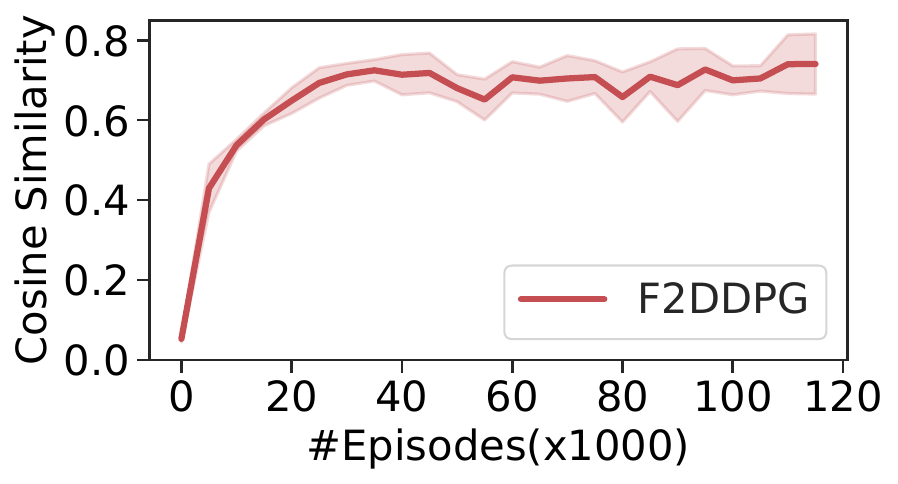}
  \captionsetup{labelformat=empty}
  \caption{b) Cosine similarity between actual actions and biases}
\end{minipage}
\caption{Results in 5 vs. 3 predator-prey.}
\label{fig:results_predator_prey}
\end{figure}

As shown in Figure \ref{fig:results_env} (c), F2DDPG outperforms other baseline algorithms with higher rewards. The reason why F2DDPG achieves higher rewards is hypothesized by investigating the reward lost by the green prey shown in Figure \ref{fig:results_predator_prey} (a). This figure compares the reward achieved by the green prey (trained with MADDPG) when playing against the predators trained with F2DDPG and baseline algorithms; when the reward of the prey is lower, it is more likely to be captured by the predators. As shown in the figure, the reward of the green prey captured by the predators trained with F2DDPG is lower than that of the prey captured by the predators trained with other baseline algorithms. This indicates that the high rewards of the predators trained with F2DDPG (shown in Figure \ref{fig:results_env} (c)) are the results of capturing the green prey more frequently. Thus, it can be concluded that the predators trained with F2DDPG are more capable of cooperating strategically to capture the most rewardable, but fast, prey than the predators trained with other baseline algorithms.

We also verify that the infused bias induces desirable behaviors from the agents. Figure \ref{fig:results_predator_prey} (b) shows how the cosine similarity between the actual action $\mathbf{a}^A_{-i}$ and bias $\nabla_{\mathbf{a}^A_{-i}}Q_i^{\mu}$ in the biased action in Equation \ref{eq:F2DDPG_bias} varies as the F2DDPG training proceeds. The cosine similarity between $\mathbf{a}^A_{-i}$ and $\nabla_{\mathbf{a}^A_{-i}}Q_i^{\mu}$ is calculated as $\frac{\langle a^A_{-i}, \nabla_{a^A_{-i}}Q_i^{\mu} \rangle}{\lVert a^A_{-i} \rVert_2 \lVert \nabla_{a^A_{-i}}Q_i^{\mu} \rVert_2}$ for all agents in $A(i)$, where $\langle\cdot,\cdot\rangle$ represents the inner product operator. The similarity is used to judge the similarity of directions of the actual and biased actions. The similarity ranges from -1, indicating exactly opposite, to 1, indicating exactly the same, and 0 indicating orthogonality or decorrelation. As shown in Figure \ref{fig:results_predator_prey} (b), the similarity increases from 0 to close to 1 as the training proceeds, indicating that the actual and biased actions (desirable actions designed by the biases) become more similar as the training proceeds. Therefore, the agents eventually execute the actions designed by the biases, and accordingly, the biases between the actual and biased actions vanish.

\setcounter{figure}{5}
\begin{figure}[h!]
\begin{minipage}[c]{.18\textwidth}
  \centering
  \includegraphics[width=0.9\textwidth]{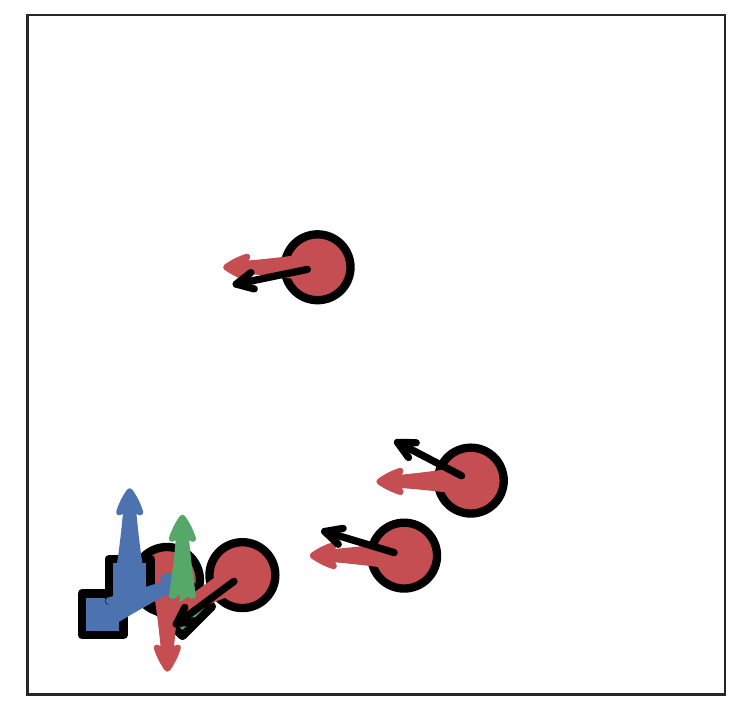}
\end{minipage}
\hspace{0.0cm}
\begin{minipage}[c]{.18\textwidth}
  \centering
  \includegraphics[width=0.9\textwidth]{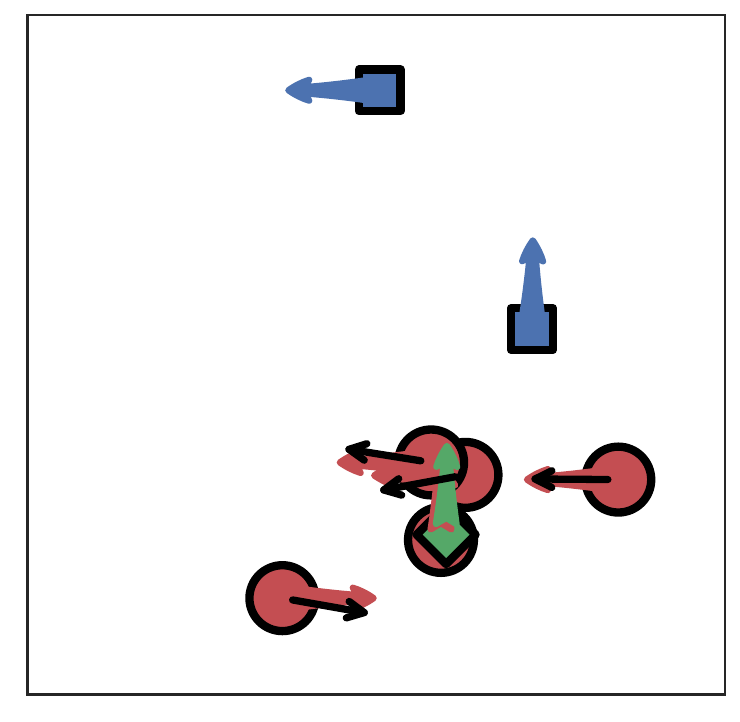}
\end{minipage}

\begin{minipage}[c]{.18\textwidth}
  \centering
  \includegraphics[width=0.9\textwidth]{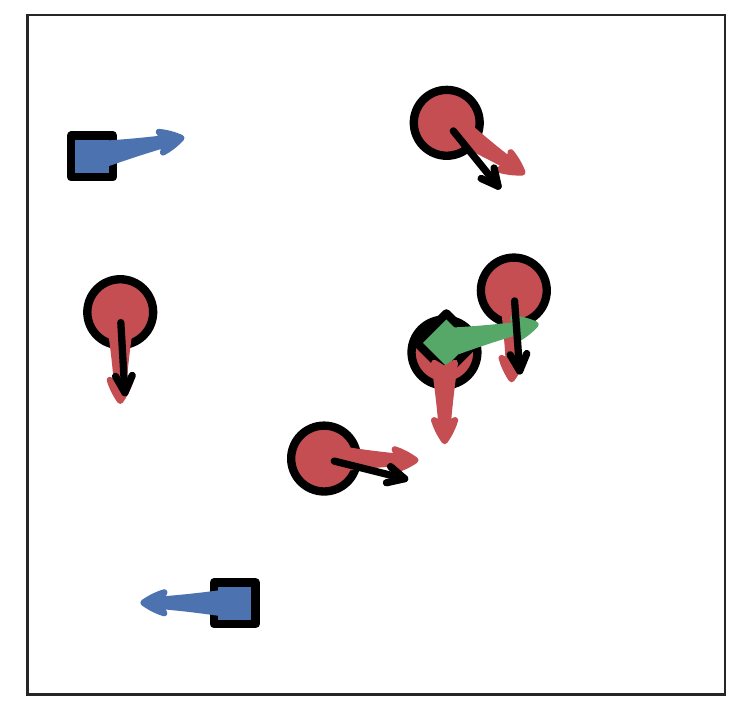}
\end{minipage}
\hspace{0.0cm}
\begin{minipage}[c]{.18\textwidth}
  \centering
  \includegraphics[width=0.9\textwidth]{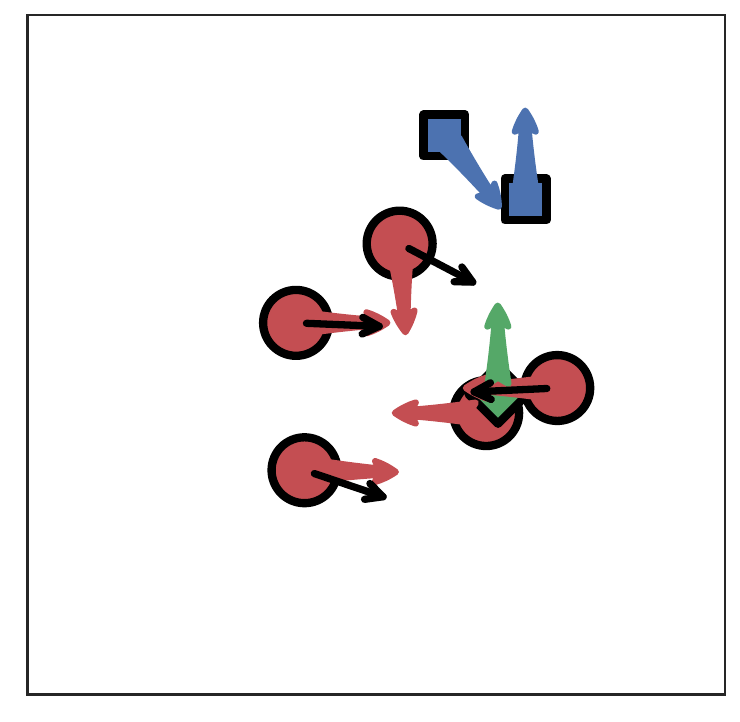}
\end{minipage}
\caption{Actual actions (red arrows) and biases (black arrows) for the predator capturing the green prey after the policies are learned with F2DDPG.}
\label{fig:random_position_action_learned}
\end{figure}

To consider a particular fixed state where a centered predator captures the prey, we investigate how the centered predator's trained policy induces other agents' (predators') actions. In Figure~\ref{fig:fixed_position_action}, the figures on the left show the other agents' biased joint action (black) induced at an early training stage, and those on the right show how the other agents' actual joint action (red) concurrently and coherently alter as the training proceeds. Noticeable observations are that (1) the other agents are jointly heading toward the prey, which is a strategic movement to capture the prey, and (2) the actual (red) and biased (black) actions become extremely similar, which demonstrates that the biases vanish as the training proceeds, and the agents actually behave as intended by the biases.

In addition, we execute the policies trained with F2DDPG in the predator-prey game with a random state. Figure \ref{fig:random_position_action_learned} shows four snapshots of the predator-prey game with a random state after the policies are trained with F2DDPG. In the figure, the predators tend to gather toward the predator capturing the green prey and attempt to capture it together. The differences between the actual (red) and biased (black) actions are negligible, meaning that the converged policies no longer carry the biases.

Table \ref{table:results_predator-prey} compares the fraction between the number of successful episodes and the total number of testing episodes (100 episodes). We define two types of success: $N_c\geq1$ is the case in which the predators capture the green prey at least once and $N_c\geq3$ is the case in which the predators capture the green prey at least three times. To validate the scalability of the proposed algorithm, we compare these performance measures for F2DDPG and other baseline algorithms for different sizes of predator-prey games. In 3 vs. 1, the performances are not significantly differentiated because the number of predators is insufficient to capture the faster green prey, even if the three predators cooperate. However, as the number of predators increases, the predators capture the green prey through cooperation, and F2DDPG outperforms other baseline algorithms in 5 vs. 3, as shown in the table. In addition, F2DDPG outperforms other baseline algorithms in 7 vs. 3. Thus, it can be concluded that the proposed method improves the MARL training, even in cases with many agents, by utilizing biased action information.

\subsection{Covert Communication}
Covert communication is a mixed cooperative-competitive environment, as shown in Figure \ref{fig:fig_illustrations} (d), with two cooperative agents, a speaker and listener (red circles), and an adversary (green circle). The speaker must encode a message as a communication vector using a randomly generated key to output the communication vector. The listener must reconstruct the communication vector into the message using the key. However, the adversary also observes the communication vector and attempts to reconstruct the communication vector without the key. The speaker and listener are rewarded based on the listener's reconstruction and penalized based on the adversary's reconstruction. The adversary is rewarded based on its reconstruction. Therefore, the speaker must encrypt the message as the communication vector such that the adversary cannot decrypt the communication vector, and the listener must decrypt the communication vector as the message. In the covert communication, the adversary is trained with MADDPG for comparison, while the speaker and listener are trained with F2DDPG and other baseline algorithms.

As shown in Figure \ref{fig:results_env} (d), F2DDPG outperforms other baseline algorithms with higher rewards. At the early stage of training, M3DDPG, Random Sign, and F2DDPG outperform MADDPG and All Plus. However, as the training proceeds and the adversary becomes intelligent, the rewards of M3DDPG and Random Sign decrease, while F2DDPG still maintains high rewards.

\section{Conclusions}
We proposed F2DDPG, an algorithm that boosts MARL training using biased action information of other agents based on a friend-or-foe concept. Empirically, we demonstrated that F2DDPG outperforms existing algorithms in several mixed cooperative-competitive environments. We also demonstrated that F2DDPG learns the agents' policies such that their actions become similar to the biased actions and that the biases decrease as the learning proceeds.  

%%% The following command should be issued somewhere in the first column 
%%% of the final page of your paper.
\balance

%%% The next two lines define, first, the bibliography style to be 
%%% applied, and, second, the bibliography file to be used.

\bibliographystyle{ACM-Reference-Format} 
% \bibliography{sample}

\newpage

\section*{Supplementary Material}

\setcounter{table}{1}
\section*{Details about Environments}
\begin{table}[h!]
  \caption{Classification of the experimental environments.}
  \label{table:classification_env}
  \centering
    \begin{tabular}{cccc}
        \toprule
        \multicolumn{1}{c}{Environment} & Cooperative? & Mixed? & Communication?  \\
        \midrule
        \multicolumn{1}{c}{Cooperative Navi.} & \checkmark & &\\
        \multicolumn{1}{c}{Cooperative Comm.} & \checkmark & &\checkmark \\
        \multicolumn{1}{c}{Predator-Prey} & &\checkmark& \\
        \multicolumn{1}{c}{Covert Comm.} & &\checkmark&\checkmark\\
        \bottomrule
    \end{tabular}
\end{table}

Table \ref{table:classification_env} categorizes the experimental environments into a cooperative environment and a mixed cooperative-competitive environment and indicates whether the environments need communication between agents. We assume that the agents in the environments observe the relative positions and velocities of all agents. The positions of the agents and landmarks in all the environments are randomly initialized within $[-1,1]^2$ for every episode.

\subsection*{Cooperative Navigation}
Cooperative navigation ($N=3$) has three agents and three landmarks. 

\subsection*{Cooperative Communication}
Cooperative communication ($N=2$) has two agents, a speaker and listener, and three landmarks. The speaker outputs a communication vector, which is a three-sized tensor, at each timestep, and the listener observes the communication vector.

\subsection*{Predator-Prey}
Predator-prey ($N=4,8,10$) has predator agents and prey agents. The environment imposes a penalty on prey when the prey go beyond the boundary of the environment.

\subsection*{Covert Communication}
Covert communication ($N=3$) has three agents, a speaker, listener, and adversary. The speaker observes a message vector and a key vector, which are randomly generated, and outputs a communication vector. The listener observes the key vector and the communication vector. The adversary observes only the communication vector. The message, key, and communication vectors are four-sized tensors.

\section*{Hyper-Parameters for Experiments}
% In this study, all the performances are obtained by the trained policies with four different random seeds.
We use 120,000 training episodes with 25 timesteps (total 3 million timesteps) for training the proposed and other baseline algorithms in all the environments. All codes used in the experiments will be released.

\subsection*{Hyper-Parameters of F2DDPG}
The hyper-parameters of F2DDPG used in the experiments are summarized in Table \ref{table:table_F2DDPG}. The output layer of the policy network of F2DDPG for the environments provides the action as a five-sized tensor for hold, right, left, up, and down. 
\begin{table}[b!]
  \caption{Hyper-parameters of F2DDPG.}
  \label{table:table_F2DDPG}
  \centering
    \begin{tabular}{cc}
        \toprule
        \multicolumn{1}{c}{F2DDPG Hyper-Parameter} &  \\
        \midrule
        \multicolumn{1}{c}{\# Policy network MLP units} &(64, 64) \\
        \multicolumn{1}{c}{\# $Q$-network MLP units} &(64, 64) \\
        \multicolumn{1}{c}{Network parameter initialization} &Xavier uniform \\
        \multicolumn{1}{c}{Nonlinear activation} &ReLU \\
        \multicolumn{1}{c}{Policy network learning rate} &$10^{-2}$ \\
        \multicolumn{1}{c}{$Q$-network learning rate} &$10^{-2}$ \\
        \multicolumn{1}{c}{$\tau$ for updating target networks} &$10^{-2}$ \\
        \multicolumn{1}{c}{$\gamma$} &0.95 \\
        \multicolumn{1}{c}{Replay buffer size} &$10^{6}$ \\
        \multicolumn{1}{c}{Mini-batch size} &1024 \\
        \multicolumn{1}{c}{Optimizer} &Adam \\
        \multicolumn{1}{c}{$\delta^A$} &$10^{-5}$ \\
        \multicolumn{1}{c}{$\delta^E$} &$10^{-3}$ \\
        \bottomrule
    \end{tabular}
\end{table}

The hyper-parameters in the table are also used for the All Plus and Random Sign algorithms in the experiments. For MADDPG and M3DDPG, the hyper-parameters reported to have the highest performance in previous studies are used in the experiments and similar to the hyper-parameters of F2DDPG.

% \balance

%%%%%%%%%%%%%%%%%%%%%%%%%%%%%%%%%%%%%%%%%%%%%%%%%%%%%%%%%%%%%%%%%%%%%%%%

\end{document}